\documentclass[conference,a4paper]{IEEEtran}
\IEEEoverridecommandlockouts

\usepackage{cite}
\usepackage{amsmath,amssymb,amsfonts}
\usepackage[ruled,vlined,linesnumbered]{algorithm2e}
\usepackage{graphicx}
\usepackage{textcomp}
\usepackage{xcolor}
\usepackage{booktabs}
\usepackage{multirow}
\usepackage{url}
\usepackage{tikz}
\usetikzlibrary{arrows.meta,positioning,fit,backgrounds,calc}
\usepackage[colorlinks=true,linkcolor=blue,citecolor=blue,urlcolor=blue]{hyperref}

\def\BibTeX{{\rm B\kern-.05em{\sc i\kern-.025em b}\kern-.08em
    T\kern-.1667em\lower.7ex\hbox{E}\kern-.125emX}}

\newcommand{\Dcnn}[1]{D^{\mathtt{conv}}_{#1}}
\newcommand{\Dmlp}[1]{D^{\mathtt{fc}}_{#1}}
\SetKwFunction{Forward}{Forward}
\SetKwProg{Fn}{Function}{:}{}

\begin{document}

\setlength{\abovedisplayskip}{3.0pt}
\setlength{\belowdisplayskip}{3.0pt}
    
\title{Energy-Efficient Spiking Neural Networks for Sensing-Aided Beam Prediction\vspace{-0.5cm}}

\author{
	\IEEEauthorblockN{Nhan~Thanh~Nguyen\IEEEauthorrefmark{1},
		Phuong~Nam~Tran\IEEEauthorrefmark{1},  
		Mengyuan~Ma\IEEEauthorrefmark{1},
		Huy T. Nguyen\IEEEauthorrefmark{2}, Ahmed Alkhateeb\IEEEauthorrefmark{3}, Markku Juntti\IEEEauthorrefmark{1}}
	\IEEEauthorblockA{
		\IEEEauthorrefmark{1}Centre for Wireless Communications, University of Oulu, P.O.Box 4500, FI-90014, Finland, \\
		\IEEEauthorrefmark{2}Faculty of Information Technology, Van Lang School of Technology, Van Lang University, Ho Chi Minh City, Vietnam, \\
		\IEEEauthorrefmark{3}School of Electrical, Computer, and Energy Engineering, Arizona State University, AZ, USA,\\
		E-mails: \{nhan.nguyen, phuong.tran, mengyuan.ma, markku.juntti\}@oulu.fi, huy.nt@vlu.edu.vn, alkhateeb@asu.edu\vspace{-0.5cm}
	}
}

\maketitle

\begin{abstract}
	Sensing-aided beam prediction can reduce the beam-training overhead of millimeter-wave (mmWave) systems by leveraging environmental information from complementary modalities such as vision and position. However, the resulting sensing and inference pipeline can be energy-intensive, motivating the use of energy-efficient spiking neural networks (SNNs) as an alternative to conventional artificial neural networks (ANNs). In this work, we develop SNN architectures for beam prediction and systematically compare them with ANNs using position-only, vision-only, and multimodal inputs. Our results show that compact ANNs consistently outperform SNNs for position-only sensing. In contrast, for vision and multimodal sensing, SNNs achieve higher accuracy than ANNs at similar energy costs and require up to an order of magnitude less energy at a comparable accuracy. For example, the multimodal SNN achieves $58.08\%$ top-1 accuracy, outperforming the closest-energy ANN by $18.11$ percentage points. We further show that SNN accuracy is limited by the binary communication between layers rather than by model size. These results demonstrate the potential of SNNs for energy-efficient sensing-aided beam prediction while highlighting their accuracy limitation.
\end{abstract}

\section{Introduction}
\label{sec:intro}

High-frequency wireless communication systems, such as millimeter-wave (mmWave) systems, achieve high data rates by using large antenna arrays to form narrow and directional beams. However, conventional codebook-based beam training typically requires sweeping over a large number of candidate beam pairs, resulting in substantial training overhead, especially for highly mobile users. To reduce this overhead, recent studies have explored side information from sensors already available in the environment, including cameras \cite{charan2022vision, ma2026knowledge, nguyen2026knowledge}, GNSS receivers \cite{morais2023position, charan2022vision}, LiDAR \cite{jiang2023lidar}, and radar \cite{demirhan2022radar,Luo2023millimeter}, to predict the optimal beam and avoid exhaustive beam sweeping. In particular, Charan \textit{et al.}~\cite{charan2022vision} showed using the real-world DeepSense~6G dataset~\cite{alkhateeb2023deepsense} that combining camera images with GNSS position can outperform either modality alone, as the two modalities provide complementary information. This motivated various studies on multimodal sensing-aided beam prediction/tracking~\cite{zakeri2026aoi, ma2026knowledge, hassam2026dynamic, cui2024sensing, zhang2025multimodal, nguyen2026knowledge}. However, existing beam prediction methods have mainly relied on conventional artificial neural networks (ANNs). While ANNs are effective, their reliance on dense multiply-accumulate (MAC) operations can lead to high inference energy consumption, limiting their use in energy-constrained network nodes.

Spiking neural networks (SNNs) offer a promising alternative for energy-efficient inference~\cite{roy2019towards, tavanaei2019deep}. Unlike conventional ANNs, SNNs communicate using binary spikes and can exploit event-driven computation on neuromorphic hardware, such as Intel's Loihi chip~\cite{davies2018loihi}. This enables expensive MAC operations to be replaced by simpler accumulate (AC) operations. At comparable bit-widths, an AC operation has been reported to consume approximately $5\times$ less energy than a MAC operation~\cite{horowitz2014computing}. SNNs have also been explored for wireless communications tasks, including neuromorphic signal detection~\cite{skatchkovsky2021neuromorphic} and energy-constrained distributed inference~\cite{liu2024energy}. However, their use for multimodal sensing-aided mmWave beam prediction has not yet been investigated. More importantly, replacing conventional ANNs with SNNs generally involves an accuracy-efficiency tradeoff: SNNs can substantially reduce inference complexity and energy consumption, but may also incur a loss in prediction accuracy. This raises the question of how effectively SNNs can exploit sensing data to improve beam prediction accuracy while retaining their computational and energy advantages.

In this paper, we investigate the accuracy-energy tradeoff of SNN- and ANN-based sensing-aided beam prediction. We develop three SNN architectures: a spiking CNN for vision-only inputs, a spiking MLP for position-only inputs, and a multimodal architecture that combines the two branches for multimodal inputs. We then vary the neural network size and characterize the accuracy-energy tradeoff of both SNNs and ANNs, enabling direct comparison under different energy budgets. We observe that, with position-only sensing, a small ANN outperforms the SNN across the entire explored energy range. In contrast, for vision and multimodal sensing, SNNs outperform ANNs at similar energy costs and achieve comparable accuracy with up to an order of magnitude less energy. However, the comparison also reveals a limitation of SNNs: increasing the model size does not allow the SNN to reach the best accuracy achieved by the ANN. We further examine the source of this accuracy gap by separating the effects of binary spiking, model capacity, and surrogate-gradient approximation in SNNs. We find that binary firing, i.e., communication between SNN layers through binary signals, is the main factor limiting accuracy of SNNs. Overall, our results demonstrate the potential of SNNs for energy-efficient sensing-aided beam prediction while highlighting their fundamental limitation on attained accuracy.

\section{System Model and Beam Prediction Problem}
\label{sec:system}

We consider a mmWave base station (BS) equipped with an $N$-element antenna array, a camera, and a predefined beamforming codebook $\mathcal{F} = \{\mathbf{f}_1,\ldots,\mathbf{f}_M\}$, where $\mathbf{f}_m \in \mathbb{C}^{N \times 1}$, serving a single-antenna mobile user. Given the channel $\mathbf{h}_u \in \mathbb{C}^{N \times 1}$ and a beam $\mathbf{f}_m$, the optimal beam that maximizes the received power is given by
\begin{equation}
	\mathbf{f}^\star = \operatorname*{arg\,max}_{\mathbf{f} \in \mathcal{F}} \left\| \mathbf{h}_u^T \mathbf{f} \right\|_2^2.
	\label{eq:optimal-beam}
\end{equation}
Finding $\mathbf{f}^\star$ through exhaustive beam search requires significant beam-training overhead, especially in mobile scenarios. To reduce this overhead, we employ machine learning to directly predict the optimal beam from side information available without beam sweeping.

We consider RGB image and GNSS position information available at the BS for beam prediction~\cite{charan2021vision}. Let $\boldsymbol{\mathcal{X}}_u \in \mathbb{R}^{W \times H \times C}$ and $\mathbf{g}_u \in \mathbb{R}^{2}$ denote the RGB image and the normalized 2-D position of user $u$, respectively. We formulate beam selection as an $M$-class classification problem. Let $\boldsymbol{\mathcal{S}} \in \{\boldsymbol{\mathcal{X}}_u,\mathbf{g}_u,(\boldsymbol{\mathcal{X}}_u,\mathbf{g}_u)\}$ denote the observation available to a vision-only, position-only, or multimodal model, respectively. A parameterized model $f_{\boldsymbol{\Theta}}$ maps the observation $\boldsymbol{\mathcal{S}}$ to a probability distribution $\boldsymbol{\mathcal{P}} = \{p_1,\ldots,p_M\}$ over the codebook $\mathcal{F}$. The predicted beam index is then $\hat{m}=\operatorname*{arg\,max}_{m} p_m$, and the corresponding beam $\hat{\mathbf{f}} = \mathbf{f}_{\hat{m}}$ is selected. Given a labeled dataset $\boldsymbol{\mathcal{D}}=\{(\boldsymbol{\mathcal{S}}_i,\mathbf{f}^{\star}_i)\}$, where $\mathbf{f}^{\star}_i$ denotes the ground-truth optimal beam for the $i$-th sample, we aim to learn model parameters
\begin{equation}
	\boldsymbol{\Theta}^\star = \operatorname*{arg\,max}_{\boldsymbol{\Theta}} \Pr\left(\hat{\mathbf{f}} = \mathbf{f}^{\star} \mid \boldsymbol{\mathcal{S}}\right),
	\label{eq:learning-objective}
\end{equation}
that maximize the probability of predicting the optimal beam. Unlike existing approaches that rely on conventional ANNs, we investigate SNNs for this task and study their accuracy--complexity--energy tradeoff, as described next.

\section{SNNs for Sensing-Aided Beam Prediction}
\label{sec:method}

We develop three SNN architectures, namely a spiking CNN, a spiking MLP, and a hybrid architecture combining the two, for vision-, position-, and multimodal vision-position-aided beam prediction, respectively. We refer to them as \textit{VisionSNN}, \textit{PositionSNN}, and \textit{MultiSNN}, respectively. We first present the spiking building blocks, including input encoding, the leaky integrate-and-fire (LIF) neuron, and the readout layer, which are common to all the three SNN architectures.

\subsection{Fundamental SNN Architecture}
\label{sec:architecture}

\subsubsection{Input Encoding}

We use direct encoding, where the same input is presented at every timestep, i.e., $\mathbf{a}_0[t] = \mathbf{x}$, $t=1,\ldots,T$, with $T$ being the number of timesteps in the SNN. This direct input encoding method avoids the sampling noise of rate-based encoding and allows temporal dynamics to arise entirely from the spiking neurons.

\subsubsection{LIF Neuron}

At the $\ell$th SNN layer, the input $\mathbf{a}_{\ell-1}[t]$ is first transformed by applying the layer weights and bias:
\begin{equation}
	\mathbf{u}_{\ell}[t] = \mathbf{W}_{\ell} \mathbf{a}_{\ell-1}[t] + \mathbf{b}_{\ell},
	\label{eq:preact}
\end{equation}
where $\mathbf{W}_{\ell}$ and $\mathbf{b}_{\ell}$ denote the layer weights and bias, respectively. Then, $\mathbf{u}_{\ell}[t]$ is processed by the leaky LIF dynamics:
\begin{align}
	\bar{\mathbf{v}}_{\ell}[t] &= \mathbf{v}_{\ell}[t-1] - \frac{1}{\tau}\left(\mathbf{v}_{\ell}[t-1]-V^{\mathtt{res}}\right) + \mathbf{u}_{\ell}[t],
	\label{eq:lif1}\\
	\mathbf{s}_{\ell}[t] &= \Phi\left(\bar{\mathbf{v}}_{\ell}[t]-V^{\mathtt{th}}\right),
	\label{eq:lif2}\\
	\mathbf{v}_{\ell}[t] &= \left(1-\mathbf{s}_{\ell}[t]\right)\odot\bar{\mathbf{v}}_{\ell}[t] + \mathbf{s}_{\ell}[t]V^{\mathtt{res}},
	\label{eq:lif3}
\end{align}
where $\mathbf{v}_{\ell}[t]$ is the membrane potential, $\tau$ is the membrane time constant, and $V^{\mathtt{th}}$ and $V^{\mathtt{res}}$ are the firing threshold and reset potential, respectively. In~\eqref{eq:lif1}, the input is integrated while the membrane potential leaks toward $V^{\mathtt{res}}$. In \eqref{eq:lif2}, $\Phi(\cdot)$ denotes the Heaviside function, such that
\begin{align}
s_{\ell n}[t]
=
\begin{cases}
1, & \bar{v}_{\ell n}[t]\geq V^{\mathtt{th}},\\
0, & \bar{v}_{\ell n}[t]<V^{\mathtt{th}},
\end{cases}
\end{align}
i.e., the $n$th neuron generates a spike ($s_{\ell n}[t]=1$) when the pre-spike membrane potential $\bar{v}_{\ell n}[t]$ reaches or exceeds the firing threshold $V^{\mathtt{th}}$ and remains silent ($s_{\ell n}[t]=0$) otherwise.
Here, $(\cdot)_{\ell n}[t]$ denotes the $n$th element of $(\cdot)_{\ell}[t]$. In \eqref{eq:lif3}, the membrane potential is reset for the next timestep. The output of each spiking layer is its binary spike vector, $\mathbf{a}_{\ell}[t]=\mathbf{s}_{\ell}[t]$, which is propagated to the next layer. Since the inputs to subsequent layers are binary, their weighted operations can be implemented as spike-driven AC operations on event-driven neuromorphic hardware instead of dense MAC operations. This can reduce computational complexity and energy consumption compared to  inference with ANN.

\subsubsection{Readout}

The readout layer takes the last spiking layer's output $\mathbf{a}_{\mathtt{rdout}}[t]$ and produces $M$ beam logits at each timestep, which are averaged over the $T$ timesteps for the final logits:
\begin{equation}
	\mathbf{z}[t] = \mathbf{W}_{\mathtt{rdout}}\mathbf{a}_{\mathtt{rdout}}[t] + \mathbf{b}_{\mathtt{rdout}}, \quad
	\hat{\mathbf{y}} = \frac{1}{T}\sum_{t=1}^{T}\mathbf{z}[t],
	\label{eq:weighted-readout}
\end{equation}
where $\mathbf{W}_{\mathtt{rdout}}\in\mathbb{R}^{M\times\dim(\mathbf{a}_{\mathtt{rdout}}[t])}$ and $\mathbf{b}_{\mathtt{rdout}}\in\mathbb{R}^{M}$ are the readout weights and bias, so that $\mathbf{z}[t]$ and $\hat{\mathbf{y}}\in\mathbb{R}^{M}$ each have $M$ elements representing logits for $M$ candidate beams.
The predicted beam index is then obtained as:
\begin{align}
	\hat{m} = \arg\max_{m \in \{1,\ldots,M\}} \hat{y}_m.
\end{align}

\subsection{VisionSNN, PositionSNN, and MultiSNN}
\label{sec:models}

We build VisionSNN, PositionSNN, and MultiSNN based on the components described above. Specifically, we apply the weighted operation in~\eqref{eq:preact}, followed by BatchNorm and the LIF dynamics in~\eqref{eq:lif1}--\eqref{eq:lif3}, to form each spiking layer of these SNNs. Here, BatchNorm is applied element-wise to $\mathbf{u}_{\ell}[t]$ as $\mathtt{BN}(\mathbf{u}_{\ell}[t]) = \boldsymbol{\gamma}\odot\frac{\mathbf{u}_{\ell}[t]-\boldsymbol{\mu}}{\sqrt{\boldsymbol{\sigma}^2+\epsilon}}+\boldsymbol{\beta}$, where $\boldsymbol{\mu}$ and $\boldsymbol{\sigma}^2$ denote the corresponding per-channel mean and variance, $\boldsymbol{\gamma}$ and $\boldsymbol{\beta}$ are learnable scale and shift parameters, $\odot$ denotes element-wise multiplication, and $\epsilon>0$ ensures numerical stability. BatchNorm helps stabilize the activation range before the LIF neuron applies the fixed threshold $V^{\mathtt{th}}$, thereby reducing excessive or insufficient spiking when stacking multiple layers.

We refer to layers with convolutional and fully connected weighted operations as \emph{spiking convolutional blocks} and \emph{spiking fully-connected layers}, respectively. We then arrange these building blocks to process vision, position, and multimodal inputs, as illustrated in Fig.~\ref{fig:architecture} and described below.

\subsubsection{VisionSNN}

This model processes the $224\times224$ input image $\boldsymbol{\mathcal{X}}_u$ using the direct input encoding in $\mathbf{a}_0[t]=\mathbf{x}$ and produces a beam prediction through spiking layers and the non-spiking readout. We employ $L_{\mathtt{vis}}$ spiking convolutional blocks with channel widths $\Dcnn{1}\rightarrow\Dcnn{2}\rightarrow\cdots\rightarrow\Dcnn{L_{\mathtt{vis}}}$. Each block applies the weighted operation in~\eqref{eq:preact}, followed by BatchNorm and the LIF dynamics in~\eqref{eq:lif1}--\eqref{eq:lif3}. The first convolution uses stride $2$, and each subsequent block is followed by $2\times2$ max pooling. We then use a global average pooling (GAP) to produce a $\Dcnn{L_{\mathtt{vis}}}$-dimensional spiking feature, which passes through one spiking fully-connected layer of width $D^{\mathtt{vis}}_{\mathtt{rdout}}$ before the readout in~\eqref{eq:weighted-readout} to determine $\hat{m}$.

\subsubsection{PositionSNN}

PositionSNN processes the normalized 2D GNSS position $\mathbf{g}_u$ using the direct input encoding introduced in Section~\ref{sec:architecture}. We construct $L_{\mathtt{pos}}$ spiking fully-connected layers with widths $2\rightarrow\Dmlp{1}\rightarrow\cdots\rightarrow\Dmlp{L_{\mathtt{pos}}}$. Each layer applies the weighted operation in~\eqref{eq:preact}, followed by BatchNorm and the LIF dynamics in~\eqref{eq:lif1}--\eqref{eq:lif3}. The resulting $D^{\mathtt{pos}}_{\mathtt{rdout}}$-dimensional spiking feature is passed directly to the non-spiking readout in~\eqref{eq:weighted-readout}, without an additional hidden layer, where we define $D^{\mathtt{pos}}_{\mathtt{rdout}}\triangleq\Dmlp{L_{\mathtt{pos}}}$ for consistency with the notation used for VisionSNN. Thus, PositionSNN uses the same fundamental SNN components as VisionSNN. The two architectures differ only in their input modality and in the weighted operation in \eqref{eq:preact}. More specifically, VisionSNN applies a 2D convolution over the spatial dimensions of $\mathbf{a}_{\ell-1}[t]$, whereas PositionSNN applies a standard matrix-vector multiplication.

\subsubsection{MultiSNN}

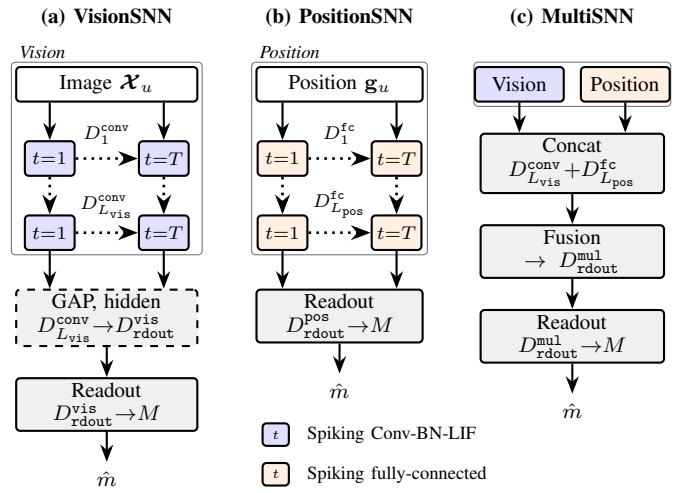
\begin{figure}[t]
	\centering
	\resizebox{\linewidth}{!}{\begin{tikzpicture}[
  node distance=4mm and 3mm,
  >=Stealth,
  thick,
  block/.style={draw, rounded corners=1.5pt, align=center, font=\footnotesize, inner sep=1.6pt, minimum height=4.6mm},
  ioblock/.style={block, fill=white},
  visblock/.style={block, fill=blue!12},
  posblock/.style={block, fill=orange!12},
  neutralblock/.style={block, fill=black!6},
  hiddenblock/.style={neutralblock, dashed},
  lbl/.style={font=\footnotesize\bfseries},
  colA/.style={text width=23mm, align=center},
  colB/.style={text width=21mm, align=center},
  colC/.style={text width=23.3mm, align=center},
  tcell/.style={text width=5.5mm, align=center, font=\scriptsize},
  anchorpt/.style={inner sep=0pt, minimum size=0pt}
]

\node[lbl] (a-title) at (0,0) {(a) VisionSNN};
\node[ioblock, colA, below=of a-title] (a-in) {Image $\boldsymbol{\mathcal{X}}_u$};
\node[anchorpt, below=7.3mm of a-in] (a-conv1-dots) {};
\node[visblock, tcell, left=4mm of a-conv1-dots] (a-conv1-t1) {$t{=}1$};
\node[visblock, tcell, right=4mm of a-conv1-dots] (a-conv1-tT) {$t{=}T$};
\draw[->, dotted, line width=1pt](a-conv1-t1) --(a-conv1-tT);
\node[font=\scriptsize\itshape, above=0.5mm of a-conv1-dots] (a-conv1-lbl) {$\Dcnn{1}$};
\node[anchorpt, below=9.6mm of a-conv1-dots] (a-conv2-dots) {};
\node[visblock, tcell, left=4mm of a-conv2-dots] (a-conv2-t1) {$t{=}1$};
\node[visblock, tcell, right=4mm of a-conv2-dots] (a-conv2-tT) {$t{=}T$};
\draw[->, dotted, line width=1pt](a-conv2-t1) --(a-conv2-tT);
\node[font=\scriptsize\itshape, above=0.5mm of a-conv2-dots] (a-conv2-lbl) {$\Dcnn{L_{\mathtt{vis}}}$};
\node[hiddenblock, colA, below=7.3mm of a-conv2-dots] (a-hid) {GAP, hidden\\$\Dcnn{L_{\mathtt{vis}}}{\to}D^{\mathtt{vis}}_{\mathtt{rdout}}$};
\node[neutralblock, colA, below=of a-hid] (a-read) {Readout $D^{\mathtt{vis}}_{\mathtt{rdout}}{\to}M$};
\node[below=of a-read, font=\footnotesize] (a-out) {$\hat{m}$};
\draw[->] (a-in.south -| a-conv1-t1.north) -- (a-conv1-t1.north);
\draw[->] (a-in.south -| a-conv1-tT.north) -- (a-conv1-tT.north);
\draw[->, dotted, line width=1pt](a-conv1-t1.south) -- (a-conv2-t1.north);
\draw[->, dotted, line width=1pt](a-conv1-tT.south) -- (a-conv2-tT.north);
\draw[->] (a-conv2-t1.south) -- (a-conv2-t1.south |- a-hid.north);
\draw[->] (a-conv2-tT.south) -- (a-conv2-tT.south |- a-hid.north);
\draw[->] (a-hid) -- (a-read);
\draw[->] (a-read) -- (a-out);
\begin{scope}[on background layer]
\node[draw=black!55, thin, rounded corners=1.5pt, inner sep=0.5mm,
      fit=(a-in)(a-conv1-lbl)(a-conv1-t1)(a-conv1-tT)(a-conv2-lbl)(a-conv2-t1)(a-conv2-tT)] (a-visframe) {};
\end{scope}
\node[font=\scriptsize\itshape, anchor=south west, inner sep=0pt] at ([xshift=1mm,yshift=0.3mm]a-visframe.north west) {Vision};

\node[lbl] (b-title) at (3.1,0) {(b) PositionSNN};
\node[ioblock, colB, below=of b-title] (b-in) {Position $\mathbf{g}_u$};
\node[anchorpt, below=7.3mm of b-in] (b-fc1-dots) {};
\node[posblock, tcell, left=4mm of b-fc1-dots] (b-fc1-t1) {$t{=}1$};
\node[posblock, tcell, right=4mm of b-fc1-dots] (b-fc1-tT) {$t{=}T$};
\draw[->, dotted, line width=1pt](b-fc1-t1) --(b-fc1-tT);
\node[font=\scriptsize\itshape, above=0.5mm of b-fc1-dots] (b-fc1-lbl) {$\Dmlp{1}$};
\node[anchorpt, below=9.6mm of b-fc1-dots] (b-fc2-dots) {};
\node[posblock, tcell, left=4mm of b-fc2-dots] (b-fc2-t1) {$t{=}1$};
\node[posblock, tcell, right=4mm of b-fc2-dots] (b-fc2-tT) {$t{=}T$};
\draw[->, dotted, line width=1pt](b-fc2-t1) --(b-fc2-tT);
\node[font=\scriptsize\itshape, above=0.5mm of b-fc2-dots] (b-fc2-lbl) {$\Dmlp{L_{\mathtt{pos}}}$};
\node[neutralblock, colB, below=7.3mm of b-fc2-dots] (b-read) {Readout\\$D^{\mathtt{pos}}_{\mathtt{rdout}}{\to}M$};
\node[below=of b-read, font=\footnotesize] (b-out) {$\hat{m}$};
\draw[->] (b-in.south -| b-fc1-t1.north) -- (b-fc1-t1.north);
\draw[->] (b-in.south -| b-fc1-tT.north) -- (b-fc1-tT.north);
\draw[->, dotted, line width=1pt](b-fc1-t1.south) -- (b-fc2-t1.north);
\draw[->, dotted, line width=1pt](b-fc1-tT.south) -- (b-fc2-tT.north);
\draw[->] (b-fc2-t1.south) -- (b-fc2-t1.south |- b-read.north);
\draw[->] (b-fc2-tT.south) -- (b-fc2-tT.south |- b-read.north);
\draw[->] (b-read) -- (b-out);
\begin{scope}[on background layer]
\node[draw=black!55, thin, rounded corners=1.5pt, inner sep=0.5mm,
      fit=(b-in)(b-fc1-lbl)(b-fc1-t1)(b-fc1-tT)(b-fc2-lbl)(b-fc2-t1)(b-fc2-tT)] (b-posframe) {};
\end{scope}
\node[font=\scriptsize\itshape, anchor=south west, inner sep=0pt] at ([xshift=1mm,yshift=0.3mm]b-posframe.north west) {Position};

\node[lbl] (c-title) at (6.2,0) {(c) MultiSNN};
\node[visblock, text width=10.7mm, align=center, below=of c-title, anchor=north east, xshift=-1mm] (c-visbr) {Vision};
\node[posblock, text width=10.7mm, align=center, right=2mm of c-visbr] (c-posbr) {Position};
\begin{scope}[on background layer]
\node[draw=black!55, thin, rounded corners=1.5pt, inner xsep=0mm, inner ysep=0.5mm, fit=(c-visbr)(c-posbr)] (c-vpframe) {};
\end{scope}
\node[neutralblock, colC, below=4mm of $(c-visbr.south)!0.5!(c-posbr.south)$] (c-concat) {Concat\\$\Dcnn{L_{\mathtt{vis}}}{+}\Dmlp{L_{\mathtt{pos}}}$};
\node[neutralblock, colC, below=of c-concat] (c-fuse) {Fusion\\$\to D^{\mathtt{mul}}_{\mathtt{rdout}}$};
\node[neutralblock, colC, below=of c-fuse] (c-read) {Readout\\$D^{\mathtt{mul}}_{\mathtt{rdout}}{\to}M$};
\node[below=of c-read, font=\footnotesize] (c-out) {$\hat{m}$};
\draw[->] (c-visbr.south) -- (c-visbr.south |- c-concat.north);
\draw[->] (c-posbr.south) -- (c-posbr.south |- c-concat.north);
\draw[->] (c-concat) -- (c-fuse);
\draw[->] (c-fuse) -- (c-read);
\draw[->] (c-read) -- (c-out);

\node[visblock, minimum width=3.5mm, minimum height=3mm, inner sep=0pt, font=\tiny] (legend-v) at ($(b-out)!0.5!(c-out)+(-24mm,-4.5mm)$) {$t$};
\node[font=\scriptsize, anchor=west] (legend-v-txt) at ([xshift=1mm]legend-v.east) {Spiking Conv-BN-LIF};
\node[posblock, minimum width=3.5mm, minimum height=3mm, inner sep=0pt, font=\tiny, below=2mm of legend-v] (legend-p) {$t$};
\node[font=\scriptsize, anchor=west] at ([xshift=1mm]legend-p.east) {Spiking fully-connected};

\end{tikzpicture}}
	\vspace{-0.5cm}
	\caption{VisionSNN, PositionSNN, and MultiSNN architectures.}
	\label{fig:architecture}
\end{figure}

As seen in Fig.~\ref{fig:architecture}(c), this architecture combines the vision and position pathways of VisionSNN and PositionSNN. Their spiking features $\mathbf{a}^{\mathtt{vis}}[t]\in\{0,1\}^{\Dcnn{L_{\mathtt{vis}}}}$ and $\mathbf{a}^{\mathtt{pos}}[t]\in\{0,1\}^{\Dmlp{L_{\mathtt{pos}}}}$, produced by the vision and position pathways at timestep $t$, are concatenated as
\begin{equation}
	\mathbf{c}[t] = \left[\mathbf{a}^{\mathtt{vis}}[t];\mathbf{a}^{\mathtt{pos}}[t]\right] \in \{0,1\}^{\Dcnn{L_{\mathtt{vis}}}+\Dmlp{L_{\mathtt{pos}}}},
	\label{eq:fusion}
\end{equation}
and passed through a single spiking fully-connected fusion layer with dimensions $(\Dcnn{L_{\mathtt{vis}}}+\Dmlp{L_{\mathtt{pos}}})\rightarrow D^{\mathtt{mul}}_{\mathtt{rdout}}$ before the non-spiking readout in~\eqref{eq:weighted-readout}. This feature-level fusion follows the commonly used early-fusion strategy in multimodal learning~\cite{baltrusaitis2019multimodal}. The fusion layer replaces the $D^{\mathtt{vis}}_{\mathtt{rdout}}$-unit hidden layer of VisionSNN. Consequently, MultiSNN has fewer parameters than VisionSNN alone.

\subsection{Training of SNN Models}
\label{sec:training}

The derivative $\Phi'(\cdot)$ of the Heaviside function in~\eqref{eq:lif2} is a Dirac delta at the threshold, i.e., zero almost everywhere and undefined at $V^{\mathtt{th}}$, so it carries no usable gradient signal for backpropagation. The SNNs are therefore trained end-to-end using backpropagation through time (BPTT) with surrogate gradients~\cite{neftci2019surrogate}. In this work, we use the arctangent surrogate to approximate $\Phi'(\cdot)$ during the backward pass,
\begin{equation}
	\Phi'\!\left(\bar{\mathbf{v}}_{\ell}[t]\right)
	\approx
	\frac{\alpha}
	{2\left[1+\left(\frac{\pi}{2}\alpha(\bar{\mathbf{v}}_{\ell}[t]-V^{\mathtt{th}})\right)^2\right]},
	\label{eq:surrogate}
\end{equation}
applied elementwise as in~\eqref{eq:lif2}, where $\alpha$ controls the slope of the surrogate gradient around the firing threshold. The surrogate gradient affects only the backward pass and does not change the binary spike generation or forward dynamics.

For each sample $(\mathbf{x}_i,m_i)$ in minibatch $\mathcal{B}$, where $m_i$ is the ground-truth beam index satisfying $\mathbf{f}_{m_i}=\mathbf{f}^{\star}_i$, the SNN is trained to minimize
\begin{equation}
	\mathcal{L}_{\mathcal{B}}
	=
	\frac{1}{|\mathcal{B}|}
	\sum_{(\mathbf{x}_i,m_i)\in\mathcal{B}}
	\mathtt{CE}(\hat{\mathbf{y}}_i,m_i),
	\label{eq:loss}
\end{equation}
where $\mathtt{CE}(\hat{\mathbf{y}}_i,m_i)
	=
	-\log
	\frac{\exp\left(\hat{y}_{i,m_i}\right)}
	{\sum_{m=1}^{M}\exp\left(\hat{y}_{i,m}\right)}$
and $\hat{y}_{i,m}$ is the final logit associated with beam $m$.

We summarize the training procedure of the SNNs in Algorithm~\ref{alg:train}. For each sample, the membrane potentials are initialized to $V^{\mathtt{res}}$, the input is presented for $T$ timesteps, and the output logits are averaged to obtain $\hat{\mathbf{y}}_i$. The minibatch loss in~\eqref{eq:loss} is then backpropagated using the gradient in \eqref{eq:surrogate}, and the model is updated accordingly, as in steps 11--17.

\begin{algorithm}[t]
	\small
	\DontPrintSemicolon
	\caption{\small Training and inference of SNN models}
	\label{alg:train}

	\SetKwInOut{Input}{Input}
	\SetKwInOut{Output}{Output}

	\Input{Training set $\mathcal{D}_{\mathtt{train}}$; parameters
		$\boldsymbol{\Theta}=\{\mathbf{W}_\ell, \mathbf{b}_\ell\}_{\ell=1}^{L+1}$; SNN model parameters $ \{T, L, \tau, V^{\mathtt{res}}, V^{\mathtt{th}}\}$; encoder $\mathcal{E}(\cdot,\cdot)$; LR schedule $\{\eta_e\}$}
	\Output{Trained $\boldsymbol{\Theta}$; predicted beam index $\hat{m}$}

	\Fn{$\mathtt{Forward}(\mathbf{x})$}{
		Initialize $\mathbf{v}_\ell[0] \leftarrow V^{\mathtt{res}}\mathbf{1}$, $\ell=1,\ldots,L$; $\hat{\mathbf{y}} \leftarrow \mathbf{0}$\;
		\For{$t \leftarrow 1$ \KwTo $T$}{
			Input encoding: $\mathbf{a}_0[t] \leftarrow \mathcal{E}(\mathbf{x}, t)$\;
			\For{$\ell \leftarrow 1$ \KwTo $L$}{
				Apply weights: $\mathbf{u}_\ell[t] \leftarrow \mathtt{BN}(\mathbf{W}_\ell \mathbf{a}_{\ell-1}[t] + \mathbf{b}_\ell)$\;
				Update $(\mathbf{v}_\ell[t], \mathbf{s}_\ell[t])$ by LIF dynamics~\eqref{eq:lif1}--\eqref{eq:lif3}\; Update $\mathbf{a}_\ell[t] \leftarrow \mathbf{s}_\ell[t]$\;
			}
			Readout to obtain $\hat{\mathbf{y}}$, based on~\eqref{eq:weighted-readout}\;
		}
		\KwRet $\hat{\mathbf{y}}$\;
	}

	\tcp{\textit{Training}}
	\For{each epoch}{
		\ForEach{minibatch $\mathcal{B} \subset \mathcal{D}_{\mathtt{train}}$}{
			\ForEach{$(\mathbf{x}_i, m_i) \in \mathcal{B}$}{
				$\hat{\mathbf{y}}_i \leftarrow \mathtt{Forward}(\mathbf{x}_i)$\;
			}
			Compute $\mathcal{L}_{\mathcal{B}}$ based on \eqref{eq:loss}\;
			Compute $\nabla_{\boldsymbol{\Theta}}\mathcal{L}_{\mathcal{B}}$  using  surrogate gradient~\eqref{eq:surrogate}\;
			$\boldsymbol{\Theta} \leftarrow \boldsymbol{\Theta} - \eta_{\text{epoch}}\,\nabla_{\boldsymbol{\Theta}}\mathcal{L}_{\mathcal{B}}$\;
		}
	}

	\tcp{\textit{Inference}}
	$\hat{\mathbf{y}} \leftarrow \mathtt{Forward}(\mathbf{x})$; $\hat{m} \leftarrow \arg\max_m \hat{y}_m$\;
\end{algorithm}

\subsection{Inference Complexity and Energy Consumption}
\label{sec:complexity}

\begin{table}[t]
	\centering
	\small
	\caption{Per-inference MAC and AC operation counts by architecture.}
	\label{tab:complexity}
	\begin{tabular}{lll}
		\toprule
		Models & $n_{\mathtt{MAC}}$ & $n_{\mathtt{AC}}$ \\
		\midrule
		VisionSNN
		& $C^{\mathtt{in}}_1C^{\mathtt{out}}_1k^2H_1^2$
		& $\bar{n}_{\mathtt{AC}}^{\mathtt{vis}}+D^{\mathtt{vis}}_{\mathtt{rdout}}M$ \\
		PositionSNN
		& $2D^{\mathtt{out}}_1$
		& $\bar{n}_{\mathtt{AC}}^{\mathtt{pos}}+D^{\mathtt{pos}}_{\mathtt{rdout}}M$ \\
		MultiSNN
		& $n_{\mathtt{MAC}}^{\mathtt{vis}}+n_{\mathtt{MAC}}^{\mathtt{pos}}$
		& $\bar{n}_{\mathtt{AC}}^{\mathtt{vis}}+\bar{n}_{\mathtt{AC}}^{\mathtt{pos}}+n_{\mathtt{AC}}^{\mathtt{mul}}$ \\
		\bottomrule
	\end{tabular}
\end{table}

The inference of the considered SNNs follows the forward pass outlined in steps 1--10 of Algorithm \ref{alg:train}. We quantify the inference cost in terms of MACs and ACs. The first layer of each branch receives dense input and is therefore counted as MACs, whereas subsequent spiking layers receive binary spikes and are counted as ACs. The readout and fusion layers have binary inputs, and thus, are also counted as ACs. 

For a convolutional layer $\ell$ (in VisionSNN and the vision branch of MultiSNN) with $C^{\mathtt{in}}_{\ell}$ input channels, $C^{\mathtt{out}}_{\ell}$ output channels, a $k\times k$ kernel, and an $H_{\ell}\times H_{\ell}$ output feature map, the operation count is $C^{\mathtt{in}}_{\ell}C^{\mathtt{out}}_{\ell}k^2H_{\ell}^2$. For a fully connected layer (in PositionSNN and the position branch in MultiSNN), the complexity is only $D^{\mathtt{in}}_{\ell}D^{\mathtt{out}}_{\ell}$, with $D^{\mathtt{in}}_{\ell}$ and $D^{\mathtt{out}}_{\ell}$ being the dimensions of the input and output vectors, respectively. Thus, with $C^{\mathtt{in}}_1=3$ for the RGB input and $D^{\mathtt{in}}_1=2$ for the position input, the MAC and AC counts of the three architectures are summarized in Table~\ref{tab:complexity}. Here, $D^{\mathtt{vis}}_{\mathtt{rdout}}$ and $D^{\mathtt{pos}}_{\mathtt{rdout}}$ are the dimensions of the last spiking layers of the vision and position branches, respectively, and $D^{\mathtt{mul}}_{\mathtt{rdout}}$ denotes the output dimension of the fusion layer. Furthermore, $\bar{n}_{\mathtt{AC}}^{\mathtt{vis}}=\sum_{\ell=2}^{L_{\mathtt{vis}}}C^{\mathtt{in}}_{\ell}C^{\mathtt{out}}_{\ell}k^2H_{\ell}^2$, $\bar{n}_{\mathtt{AC}}^{\mathtt{pos}}=\sum_{\ell=2}^{L_{\mathtt{pos}}}D^{\mathtt{in}}_{\ell}D^{\mathtt{out}}_{\ell}$, and $n_{\mathtt{AC}}^{\mathtt{mul}}
=\left(D^{\mathtt{vis}}_{\mathtt{rdout}}+D^{\mathtt{pos}}_{\mathtt{rdout}}\right)D^{\mathtt{mul}}_{\mathtt{rdout}}+D^{\mathtt{mul}}_{\mathtt{rdout}}M$ are the AC counts in the hidden layers of the VisionSNN, PositionSNN, and in the fusion and readout layers of the MultiSNN, respectively.

The inference energy of an SNN model is estimated as
\begin{equation}
	\hat{E}_{\mathtt{infer}}=n_{\mathtt{MAC}}E_{\mathtt{MAC}}+n_{\mathtt{SOP}}E_{\mathtt{AC}},
	\label{eq:energy}
\end{equation}
where $n_{\mathtt{MAC}}$ is given in Table \ref{tab:complexity}, and $E_{\mathtt{MAC}}$ and $E_{\mathtt{AC}}$ denote the energy costs of one MAC and one AC operation, respectively. Here, $n_{\mathtt{SOP}}$ denotes the number of effective synaptic operations (SOPs) incurred by the weighted input computation in \eqref{eq:preact}.\footnote{Following common practice in the SNN literature, we neglect the computational cost of the LIF dynamics in \eqref{eq:lif1}--\eqref{eq:lif3}, analogous to neglecting the activation-function cost in ANNs.} Let $\mathcal{L}_{\mathtt{bin}}$ denote the set of layers with binary inputs, i.e., layers $\ell$ for which $\mathbf{a}_{\ell-1}[t] \in \{0,1\}^{D_{\mathtt{in}}^{\ell}}$. The total number of SOPs over the $T$ timesteps is given as
\begin{equation}
	n_{\mathtt{SOP}}=\sum_{\ell \in \mathcal{L}_{\mathtt{bin}}} \tilde{D}_{\mathtt{out}}^{\ell}\sum_{t=1}^{T} \left\|\mathbf{a}_{\ell-1}[t]\right\|_0,
	\label{eq:total-sop-count}
\end{equation}
where  $\tilde{D}_{\mathtt{out}}^{\ell}$ denotes the average number of ACs performed by one input element, and $\|\cdot\|_0$ counts the number of nonzero elements of a vector.

From \eqref{eq:energy}, \eqref{eq:total-sop-count}, the inference energy reduction of an SNN compared with an ANN is mainly attributed to two factors. First, an AC operation consumes less energy than a MAC operation, i.e., $E_{\mathtt{MAC}} > E_{\mathtt{AC}}$ \cite{horowitz2014computing}. Second, in an SNN, AC operations are performed only for nonzero input spikes, so the number of SOPs depends directly on the sparsity of the binary input vectors. We characterize this sparsity by the \textit{firing rate}. Specifically, for a binary input vector $\mathbf{a}_{\ell-1}[t]$, a firing rate close to one, i.e., $\|\mathbf{a}_{\ell-1}[t]\|_0 / D_{\mathtt{in}}^{\ell} \approx 1$, means that nearly all neurons generate spikes and hence AC operations are performed at nearly all output neurons. In contrast, a lower firing rate reduces the number of effective SOPs and thus the inference energy, as will be demonstrated in the next section.


\section{Simulation Results}
\label{sec:results}

\subsection{Simulation Setup}
\label{sec:setup}

We evaluate the proposed SNNs on Scenario~5 of the DeepSense~6G
dataset~\cite{alkhateeb2023deepsense}.
The dataset contains $2300$ samples, each consisting of an RGB
image, GNSS position, and a 64-beam power vector. Following
\cite{charan2022vision}, we downsample the codebook to $M{=}32$ beams
and assign each sample the beam corresponding to the maximum received
power. We train all the ANNs and SNNs
independently for three random seeds, all with train/validation split of 7/3, and we present the average results. Simulation parameters are summarized in
Table~\ref{tab:sim-params}. We implement the ANNs proposed in~\cite{charan2022vision}, which serve as baselines for the other considered ANNs and SNNs.  We evaluate the considered machine learning models using top-$k$ accuracy, which measures the percentage of test samples for which the ground-truth beam is included in the top $k$ predictions.

\begin{table*}[t]
	\centering
	\small
	\caption{Simulation parameters.}
	\label{tab:sim-params}
	\setlength{\tabcolsep}{3pt}
	\begin{tabular}{lp{15.5cm}}
		\toprule
        PositionSNN & {$L_{\mathtt{pos}}{=}3$; $\mathbf{d}_{\mathtt{fc}} = \zeta \times [8,16,8]$,  with scaling factors $\zeta{\in}\{$1, 2, 4, 8, 16$\}$} \\
		VisionSNN & {$L_{\mathtt{vis}}{=}4$; $\mathbf{d}_{\mathtt{conv}} = \zeta \times [2,4,8,16]$, $D^{\mathtt{vis}}_{\mathtt{rdout}}{=}16\zeta$, with $\zeta{\in}\{$1, 2, 3, 4, 8, 16, 32$\}$} \\
		MultiSNN & {Vision branch reuses $\mathbf{d}_{\mathtt{conv}}$; $D^{\mathtt{mul}}_{\mathtt{rdout}}{=}16\zeta$, with $\zeta{\in}\{$1, 1.5, 2, 4, 8, 16$\}$; position branch fixed at $\mathbf{d}_{\mathtt{fc}} = 8 \times [8,16,8]$} \\
		\midrule
        PositionANN & {$\mathbf{d}_{\mathtt{fc}} = \zeta \times [8,16,8]$,  with $\zeta{\in}\{$1, 2, 4, 8$\}$} \\
		{VisionANN} & {$\mathbf{d}_{\mathtt{conv}} = \zeta \times [8,16,32,64]$,  with $\zeta{\in}\{$1, 2, 2.5, 3$\}$} \\
		{MultiANN} & {$\mathbf{d}_{\mathtt{conv}} = \zeta \times [8,16,32,64]$, with $\zeta{\in}\{$1, 2, 2.5, 3, 4$\}$; position branch fixed at $\mathbf{d}_{\mathtt{fc}} = 8 \times [8,16,8]$} \\
		ANN base.~\cite{charan2022vision} & ResNet-50 (vision/multimodal), $[1024,1024,1024]$ MLP (position) \\
		\midrule
		Optimizer & Adam, weight decay $10^{-4}$, peak LR $10^{-3}$, linear warmup + cosine decay, $\alpha = 2$ \\
		SNN dynamics & $T{=}8$ timesteps, $\tau{=}2.0$, $V^{\mathtt{th}}{=}1.0$, $V^{\mathtt{res}}{=}0.0$ \\
		Training length & $30$ epochs, batch size = $8$ (Vision/MultiSNN); $60$ epochs, batch size = $32$ (PositionSNN); use early stopping \\
		Energy model & $E_{\mathtt{MAC}}{=}4.6$\,pJ, $E_{\mathtt{AC}}{=}0.9$\,pJ~\cite{horowitz2014computing} \\
		\bottomrule
	\end{tabular}
\end{table*}



\subsection{Accuracy-Energy Tradeoff}
\label{sec:accuracy}

\begin{figure*}[t]
	\centering
	\vspace{-0.25cm}\includegraphics[width=\linewidth]{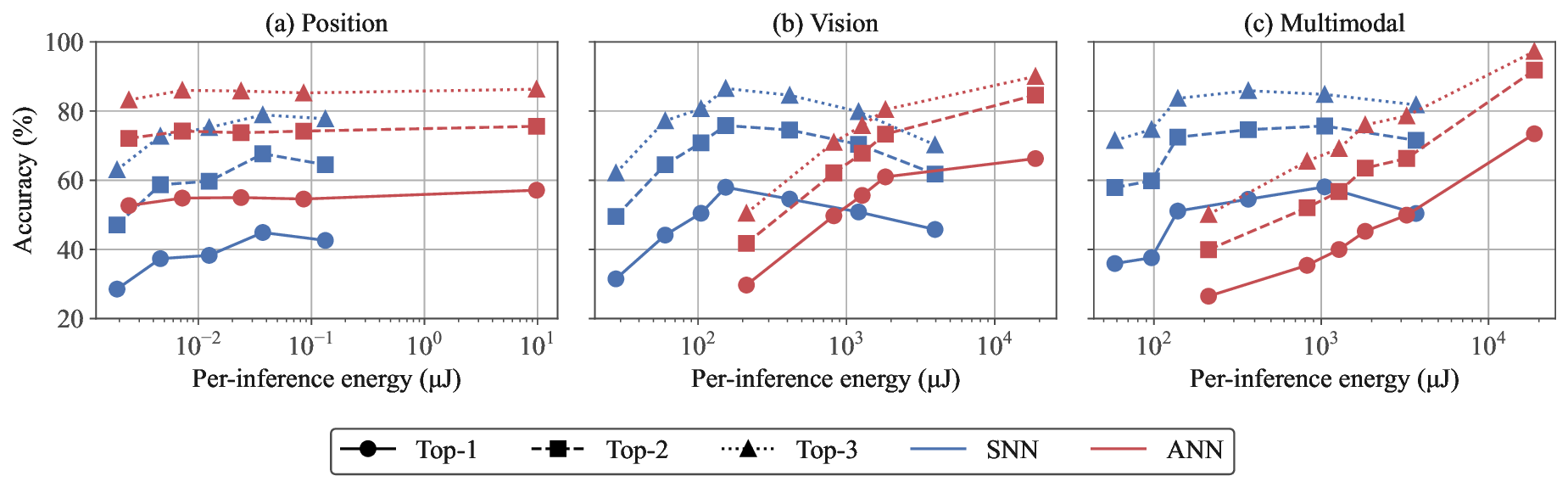}
	\caption{Beam prediction accuracy vs.\ estimated per-inference energy of the considered SNN and ANN architectures. Each ANN curve's rightmost point (with highest accuracy and energy) is obtained by the baseline models presented in~\cite{charan2022vision}.}\vspace{-0.5cm}

	\label{fig:pareto}
\end{figure*}

To compare the accuracy-energy tradeoff of SNNs and ANNs, we vary the network width of the {VisionSNN}/ANN, {PositionSNN/ANN}, and {MultiSNN/ANN} models using the scaling factor $\zeta$ specified in Table~\ref{tab:sim-params}. Each scaled model is trained and tested independently, providing a range of models with different accuracy-energy operating points, as shown in Fig.~\ref{fig:pareto}. Furthermore, we show in Table~\ref{tab:energy-sweep} the best SNN operating point and the closest ANN configuration that uses no less energy, allowing us to compare the accuracy at similar energy levels.

\begin{table}[t]\small
\centering
\caption{SNN at its best operating point vs.\ the cheapest ANN that has reasonable accuracy and uses no less energy.}
\label{tab:energy-sweep}
\setlength{\tabcolsep}{3pt}
\begin{tabular}{lrrrrr}
\toprule
 & \multicolumn{2}{c}{SNN} & \multicolumn{2}{c}{ANN} & \\
\cmidrule(lr){2-3} \cmidrule(lr){4-5}
Modality & $\mu$J & Top-1 (\%) & $\mu$J & Top-1 (\%) & $\Delta$ (pp) \\
\midrule
Position & 0.037 & 44.89 & 0.085 & 54.56 & -9.68 \\
Vision & 153.522 & 57.97 & {822.700} & {49.72} & {+8.25} \\
Multimodal & 1048.579 & 58.08 & 1277.701 & 39.97 & +18.11 \\
\bottomrule
\end{tabular}\vspace{-0.05cm}
\end{table}

\textit{Position SNN/ANN:} From Fig.~\ref{fig:pareto}(a), we observe that the ANN outperforms the SNN across the entire considered energy range. As shown in Table~\ref{tab:energy-sweep}, the ANN achieves $54.56\%$ top-1 accuracy at $0.085\,\mu$J, while the SNN achieves $44.89\%$ at $0.037\,\mu$J. More importantly, the ANN can achieve higher accuracy than the SNN at a lower energy level. This trend differs from those observed for the vision and multimodal cases discussed below. The position input has only two dimensions, resulting in relatively small position-only networks. Consequently, the energy savings from spike-driven computation are less significant than those for the larger vision and multimodal networks. Thus, SNNs do not necessarily provide an energy advantage when the underlying network is small.

\textit{Vision and Multimodal SNN/ANN:} From Fig.~\ref{fig:pareto}(b) and Fig.~\ref{fig:pareto}(c), we observe a different trend for the vision and multimodal models. Over most of the explored energy range, the SNNs achieve higher accuracy than the ANNs at comparable energy levels. This is also seen in Table~\ref{tab:energy-sweep}, where the SNN improves top-1 accuracy over the closest ANN operating point by $\Delta={8.25}$ percentage points for vision and $\Delta=18.11$ percentage points for multimodal sensing. Fig.~\ref{fig:pareto} further shows that the ANN requires substantially more energy to reach an accuracy comparable to that of the SNN. In particular, achieving similar accuracy requires roughly an order of magnitude more energy for the ANN. However, sufficiently large ANNs can eventually achieve higher accuracy than the SNNs, but only at a substantially higher energy cost. These results demonstrate that the energy advantage of SNNs becomes more significant for large vision and multimodal neural networks, where spike-driven computation can substantially reduce the inference energy while maintaining competitive accuracy.

\subsection{SNN Firing Rate}

\begin{figure}
    \centering
    \includegraphics[width=0.95\linewidth]{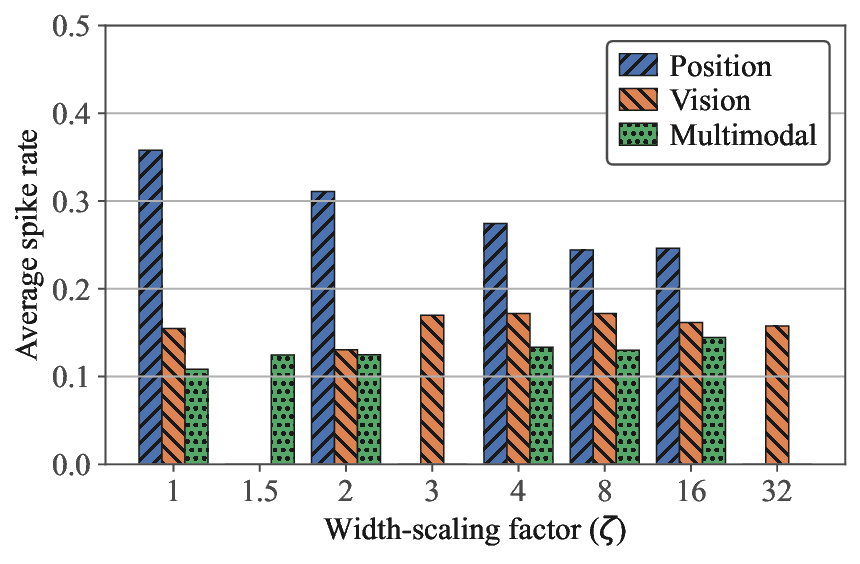}
    \caption{Firing rate of the considered SNNs.}
    \label{fig:spike-rate-pareto}
\end{figure}

To examine the role of firing rate in the energy reduction observed in Fig.~\ref{fig:pareto}, Fig.~\ref{fig:spike-rate-pareto} shows the mean firing rates of the SNN models as a function of the width scaling factor across the considered width sweep. It is observed that the mean firing rate remains below $0.5$ across the entire range of model widths. Specifically, it ranges from $0.244$ to $0.358$ for position, from $0.131$ to $0.172$ for vision, and from $0.108$ to $0.144$ for multimodal models. Moreover, the firing rate varies only moderately across the width sweep. Therefore, the energy reduction observed in Fig.~\ref{fig:pareto} is not specific to a particular model width, but remains consistently supported by sparse spike-driven computation across the considered model sizes.

\subsection{SNN Performance Limitation}
\label{sec:diagnostic}

Another observation from Fig.~\ref{fig:pareto} is that increasing the SNN width does not continuously improve accuracy, particularly for the vision and multimodal models. To investigate this limitation, we consider three VisionSNN variants with the same architecture, summarized in Table~\ref{tab:vision-diagnostic}: (1) VisionSNN-SigmoidGrad retains binary spikes but replaces the arctangent surrogate gradient in~\eqref{eq:surrogate} with the exact sigmoid gradient, (2) VisionSNN-Sigmoid replaces binary spikes with continuous sigmoid outputs, and (3) VisionSNN-ReLU replaces the LIF neuron with ReLU and removes the $T$-step unrolling. {Here, VisionSNN-Sigmoid improves top-1 accuracy from $57.97\%$ to $67.01\%$, whereas VisionSNN-SigmoidGrad and VisionSNN-ReLU achieve only $46.47\%$ and $29.68\%$, respectively. Since VisionSNN-SigmoidGrad changes only the surrogate gradient while preserving binary firing, its degradation suggests that the surrogate gradient is not the main accuracy bottleneck. In contrast, replacing binary spikes with continuous sigmoid outputs substantially improves accuracy, indicating that binary firing is the dominant accuracy limitation of VisionSNN. However, continuous outputs require dense MACs instead of ACs, increasing the per-operation energy cost and highlighting the tradeoff in Fig.~\ref{fig:pareto}.}

\begin{table}[t]\small
\centering
\caption{Accuracy of different variants of the VisionSNN with $\zeta=4$.}
\label{tab:vision-diagnostic}
\setlength{\tabcolsep}{3pt}
\begin{tabular}{lccc}
\toprule
Variant & Top-1 (\%) & Top-2 (\%) & Top-3 (\%) \\
\midrule
VisionSNN & {57.97} & {75.81} & {86.55} \\
VisionSNN-SigmoidGrad & {46.47} & {66.18} & {78.09} \\
VisionSNN-Sigmoid & {67.01} & {82.67} & {88.08} \\
VisionSNN-ReLU & {29.68} & {41.72} & {50.45} \\
\midrule
ANN baseline (ResNet-50) & {66.26} & {84.62} & {90.03} \\
\bottomrule
\end{tabular}
\end{table}

\section{Conclusion}
\label{sec:conclusion}

In this work, we compared SNN- and ANN-based sensing-aided beam prediction using position-only, vision-only, and multimodal inputs. Rather than relying on a single architecture, we swept the network width of both families to characterize their accuracy-energy tradeoff. For vision and multimodal sensing, SNNs generally outperform ANNs at comparable energy and require roughly an order of magnitude less energy to reach similar accuracy, while position-only sensing is an exception due to its much smaller networks. We further showed that the SNN accuracy ceiling is mainly due to binary spiking rather than model capacity. Removing binary firing improves accuracy but eliminates the main energy advantage of SNNs. These results highlight the fundamental tradeoff between accuracy and energy in spike-based computation. In future works, we will focus on improving SNN performance without sacrificing its energy advantage.

\bibliographystyle{IEEEtran}
\bibliography{IEEEabrv, refs}

\end{document}